\documentclass[conference,compsoc]{IEEEtran}

\usepackage{graphicx}
\usepackage{amsmath}
\usepackage{booktabs}
\usepackage[table]{xcolor}
\renewcommand{\paragraph}[1]{\textbf{#1.}}
\usepackage{algorithm}
\usepackage{algorithmic}
\usepackage{subfigure}
\usepackage{enumitem}
\usepackage{url}
\usepackage{hyperref}

\ifCLASSOPTIONcompsoc
  \usepackage[nocompress]{cite}
\else
  \usepackage{cite}
\fi

\ifCLASSINFOpdf
\else
\fi

\begin{document}
\title{\raisebox{-0.4em}{\includegraphics[height=1.5em]{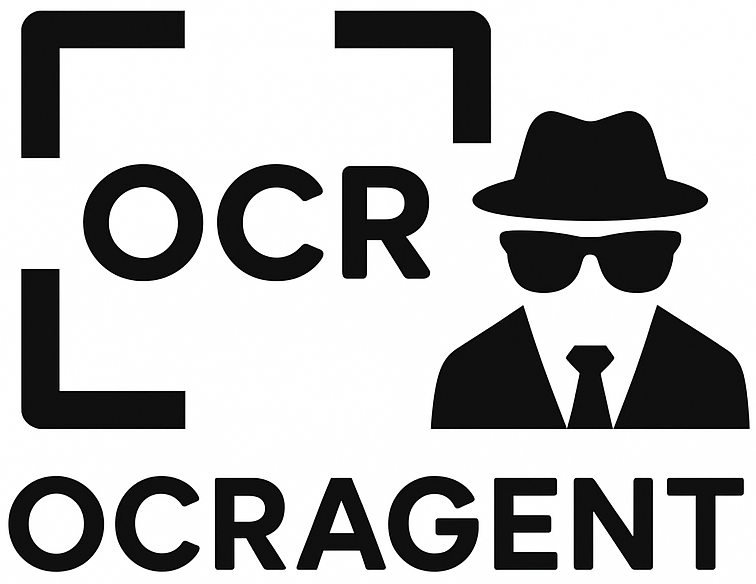}}~OCR-Agent: Agentic OCR with Capability and Memory Reflection}

\author{
    \textbf{Shimin Wen}$^{1}$\quad
    \textbf{Zeyu Zhang}$^{2*}$\quad 
    \textbf{Xingdou Bian}$^{1}$\quad 
    \textbf{Hongjie Zhu}$^{1}$\\
    \textbf{Lulu He}$^{1}$\quad
    \textbf{Layi Shama}$^{1}$\quad
    \textbf{Daji Ergu}$^{1}$\quad
    \textbf{Ying Cai}$^{3\dag}$ \vspace{0.1cm}\\
    $^1$Southwest Minzu University\quad
    $^2$AI Geeks\vspace{0.05cm}\\
    \small $^*$Project lead. $^\dag$Corresponding author: caiying@swun.edu.cn.
}

\maketitle

\begin{abstract}
Large Vision-Language Models (VLMs) have demonstrated significant potential on complex visual understanding tasks through iterative optimization methods.However, these models generally lack effective self-correction mechanisms, making it difficult for them to independently rectify cognitive biases. Consequently, during multi-turn revisions, they often fall into repetitive and ineffective attempts, failing to achieve stable improvements in answer quality.To address this issue, we propose a novel iterative self-correction framework that endows models with two key capabilities: Capability Reflection and Memory Reflection. This framework guides the model to first diagnose errors and generate a correction plan via Capability Reflection, then leverage Memory Reflection to review past attempts to avoid repetition and explore new solutions, and finally, optimize the answer through rigorous re-reasoning. Experiments on the challenging OCRBench v2 benchmark show that OCR-Agent outperforms the current open-source SOTA model InternVL3-8B by +2.0 on English and +1.2 on Chinese subsets, while achieving state-of-the-art results in Visual Understanding (79.9) and Reasoning (66.5) — surpassing even larger fine-tuned models. Our method demonstrates that structured, self-aware reflection can significantly enhance VLMs’ reasoning robustness without additional training.
Code: \url{https://github.com/AIGeeksGroup/OCR-Agent}.

\end{abstract}

\IEEEpeerreviewmaketitle

\section{Introduction}
Optical Character Recognition (OCR) constitutes a fundamental technology that bridges the visual and textual domains, aiming to extract and interpret text from images into machine-readable formats. Most recently, Large VLMs have demonstrated exceptional promise in OCR-related tasks, exhibiting strong zero-shot abilities that can be further enhanced through task-specific fine-tuning, as evidenced by models like OlmOCR.

However, directly transferring these prompting strategies to VLMs has not yielded satisfactory results \cite{zhang2024mathverse}—in some question-answering scenarios, performance may even fall short of direct answering. This is largely due to two key challenges: capability hallucination, where models propose actions beyond their executable scope (e.g., image enhancement or human proofreading), and refinement stagnation, where models get stuck in repetitive or ineffective correction loops. Current research primarily focuses on workarounds such as fine-tuning \cite{cheng2024vision} or reinforcement learning \cite{wang2025vl}, rather than addressing these core reasoning failures.This gap highlights the need for a reasoning-focused approach that enables models to self-correct within their inherent capabilities.

To this end, we demonstrate that carefully designed and properly constrained self-reflection mechanisms can enable CoT prompting to achieve sustained and stable performance improvement, effectively mitigating the limitations of standard CoT in vision-language tasks.

We propose \textbf{OCR-Agent}, a novel reflective framework designed to significantly enhance the stability and effectiveness of iterative self-correction in VLMs. Our agent architecture incorporates two key mechanisms: multi-turn \textbf{Capability Reflection}, which guides the model to diagnose errors and adaptively plan corrective actions, and \textbf{Memory Reflection}, which enables the agent to retain and leverage historical reasoning traces—avoiding redundant attempts while exploring new solution pathways. 

\begin{figure*}[t]
    \centering
    \includegraphics[width=\textwidth]{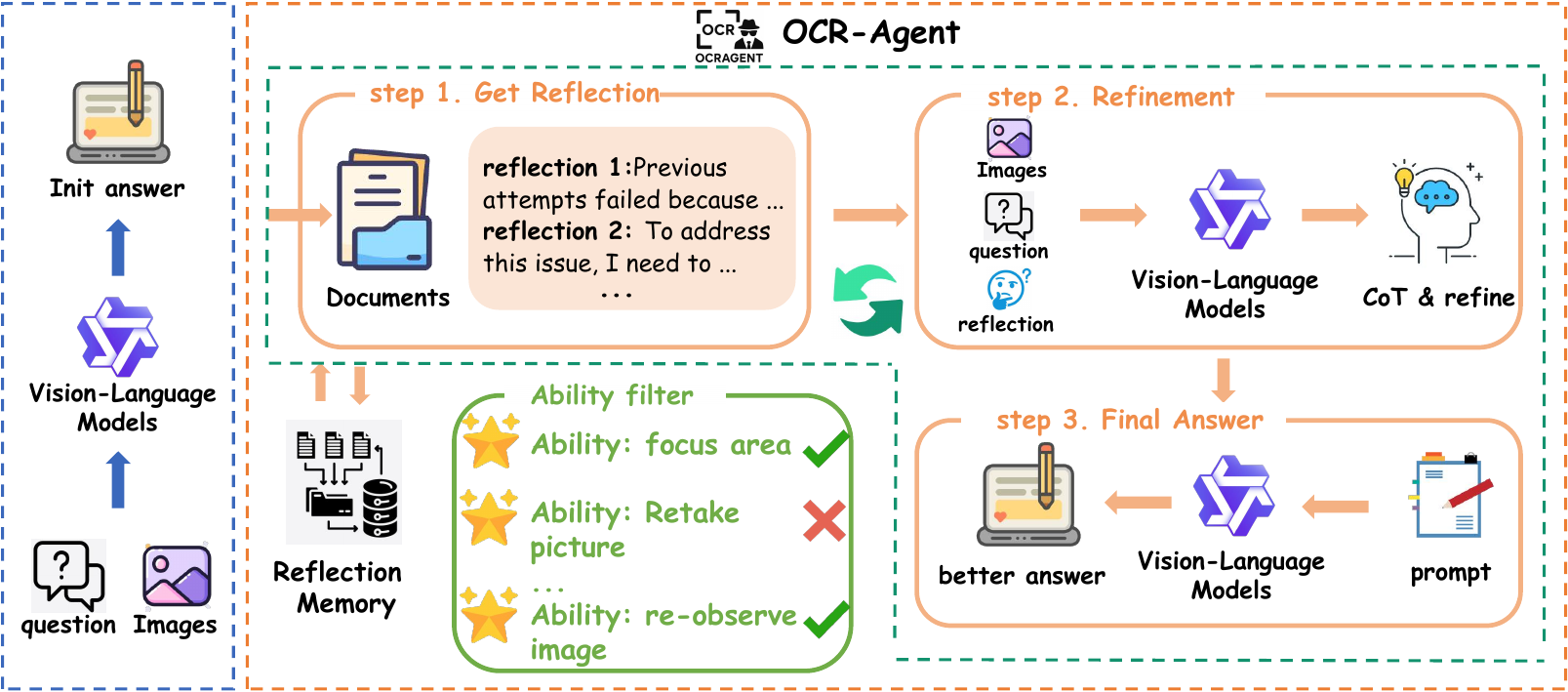}
    \caption{\textbf{Overview of OCR-Agent.}}
    \label{fig:overview}
\end{figure*}

Experimental results on the OCRBench v2 benchmark show that our method delivers substantial improvements over naive, chain-of-thought (CoT) prompting, and Self-Refine, especially in tasks demanding fine-grained visual-textual alignment and multi-step reasoning.An overview of OCR-Agent is provided in Fig~\ref{fig:overview}. 

\begin{figure}[t!]
    \centering
    \includegraphics[width=\columnwidth]{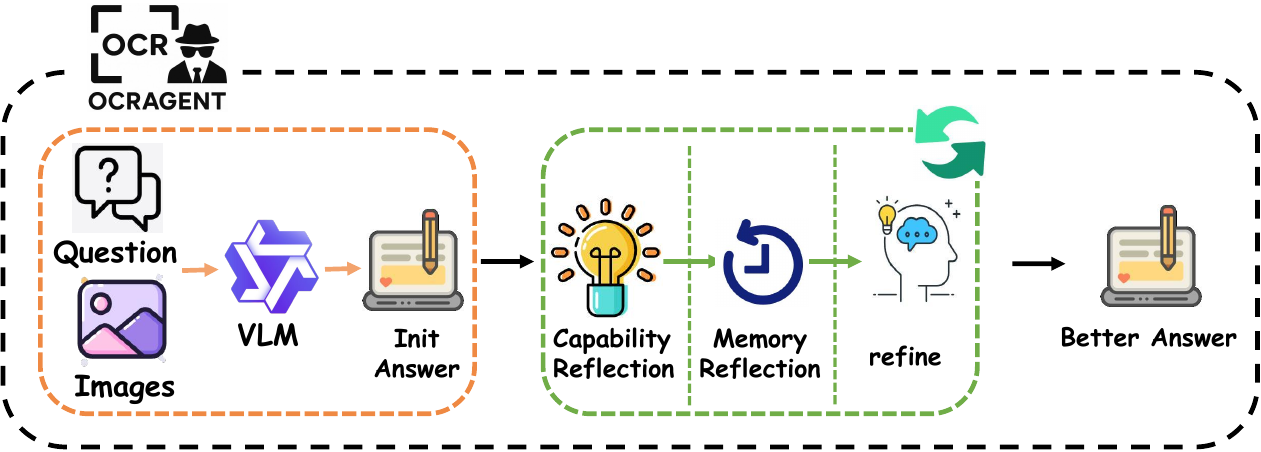}
    \caption{Overview of OCR-Agent. The model iteratively refines its answer by (1) Capability Reflection to filter infeasible actions (e.g., "enhance image"), and (2) Memory Reflection to avoid repeating past mistakes, enabling stable, training-free self-correction.}
    \label{fig:main}
\end{figure}

To summarize, our contributions are the following:
\begin{itemize}[noitemsep, topsep=0pt]  
    \item We demonstrate that specific self-reflection mechanisms can consistently and effectively enhance the performance of VLMs.
    \item We propose training-free \textbf{OCR-Agent} with two key mechanisms: \textbf{Capability Reflection} and \textbf{Memory Reflection}.
    \item We conduct experiments on the OCRBench v2, demonstrating that our method outperforms the open-source SOTA model InternVL3-8B by 2.0 points on the English and 1.2 points on the Chinese subset.
\end{itemize}

\section{Related Work}

\paragraph{Optical Character Recognition (OCR)} OCR is a key technology aimed at converting text within images into a machine-readable format. Traditional OCR methods, such as earlier versions of the Tesseract engine \cite{4376991}, typically employ a multi-stage pipeline. In the character recognition stage, these methods often rely on pattern matching (or statistical pattern recognition) to model partial features of characters, and utilize simple contextual rules or statistical associations \cite{mori1992historical}.

To overcome the limitations of traditional methods, researchers have gradually shifted towards end-to-end models based on deep learning, with approaches combining computer vision (CV) and natural language processing (NLP) becoming mainstream. For instance, the CRNN model \cite{Shi2015AnET} employs a Convolutional Neural Network (CNN) for feature extraction and combines it with a Recurrent Neural Network (RNN) for sequence transcription. The computer vision-based approach has enhanced the adaptability of text detection and instance segmentation \cite{long2018textsnake}, and has proven effective in modeling the spatial structure of entire documents \cite{katti2018chargrid}. Subsequently, Transformer-based models, such as TrOCR \cite{Li2021TrOCRTO}, leverage their powerful self-attention mechanisms to capture long-range dependencies.

In recent years, the rise of large VLMs has propelled OCR technology into a new phase of development. Studies on pretrained VLMs \cite{liu2023visual} have revealed their remarkable zero-shot OCR capabilities, with further fine-tuning leading to significant performance improvements. For example, Olmocr \cite{Poznanski2025olmOCRUT} is fine-tuned based on Alibaba's Qwen-2.5-7B-Instruct model. However, when confronted with few-shot text and real-world complexities, VLMs still face challenges in handling intricate problems and achieving practical deployment \cite{liu2024ocrbench} \cite{fu2024ocrbench}.

\paragraph{Self Reflection} Chain of Thought (CoT) \cite{Wei2022ChainOT} is one of the pioneering works that enables large language models to acquire multi-step reasoning capabilities. Through simple prompting (Zero-shot-CoT) \cite{kojima2022large} or fine-tuning \cite{kim2023cot} \cite{hsieh2023distilling}, CoT guides the model to generate a series of intermediate reasoning steps before providing a final answer. 

Building upon CoT, Self-Refine \cite{madaan2023self} introduced a more general framework for iterative improvement, and the Reflexion framework \cite{shinn2023reflexion} elevates the concept of self-reflection to a new level by structuring the language model as an agent capable of reflecting on task feedback and maintaining its own memory.

\section{Method}

\subsection{Overview}

\begin{figure*}[t]
    \centering
    \includegraphics[width=\linewidth]{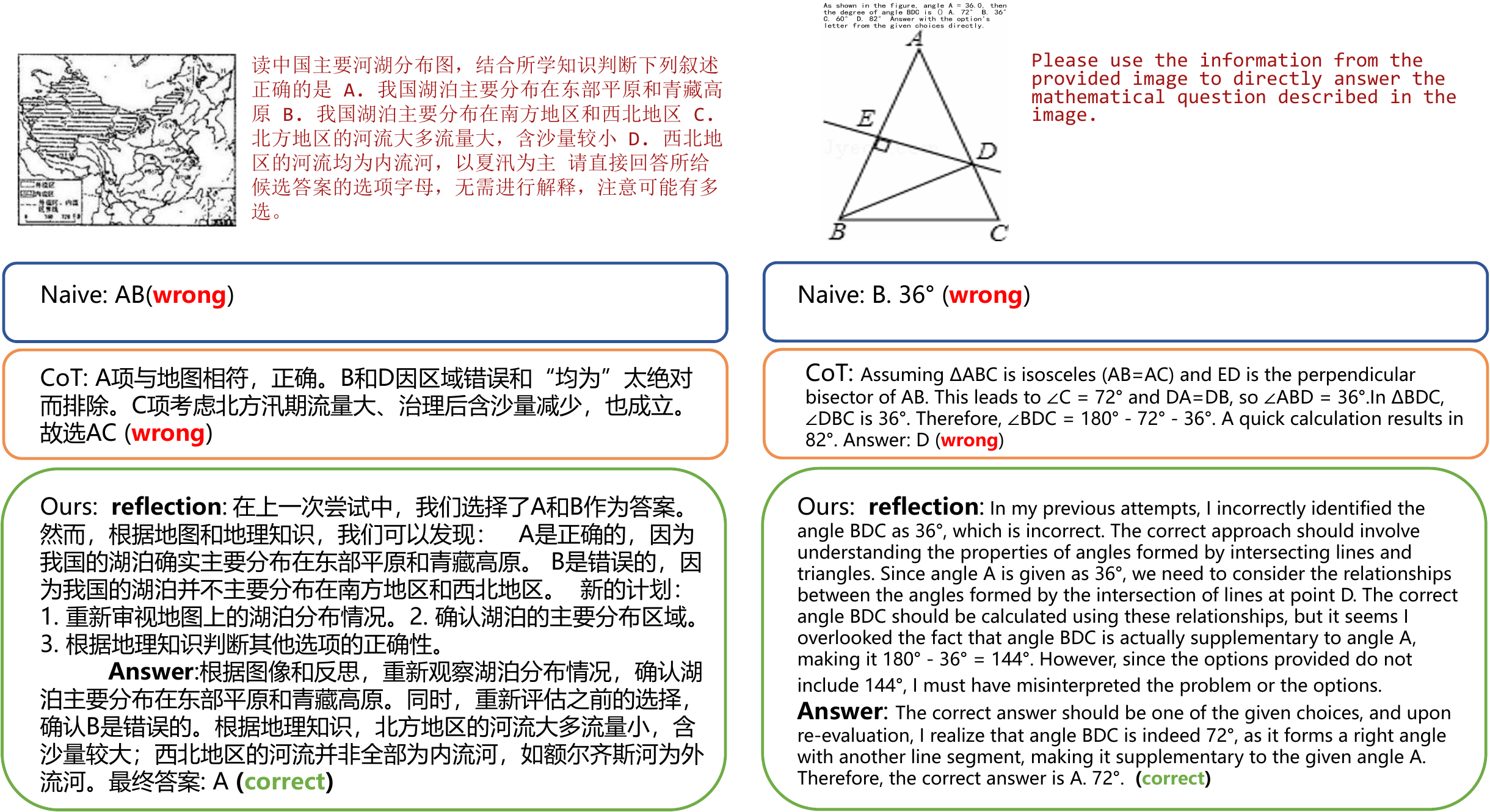}
    \caption{A visual comparison of results from the Naive, CoT, and our proposed methods. The examples in the "Ours" column represent the final output after the last round of iteration.}
    \label{fig:vis}
\end{figure*}

We propose an iterative self-reflection framework designed specifically for VLMs. This framework extends the paradigm of self-reflection from language models, moving beyond the correction of textual logic to address multi-modal errors caused by visual perception through a unique 'reflection-refinement' iterative loop. The model first performs reflection, tracing and attributing output errors to specific visual features within the image. More critically, we introduce a "capability constraint" mechanism to govern this process. This mechanism requires the model to be aware of its own capability boundaries when planning corrective steps, proactively filtering out any "capability hallucinations" it cannot execute. Subsequently, in the refinement stage, this reflection which is both visually insightful and practically feasible is converted into an internal directive that guides the model to refocus on and deeply reinterpret the key areas of the image.The framework visualization are shown in Fig~\ref{fig:vis}.
\subsection{Capability Reflection}

We proposed the capability reflection, in which the model’s post-hoc planning is aware of its own capabilities and excludes any infeasible actions. In typical multi-round self-reflection (e.g. for OCR transcription/translation), the model may suggest steps like "apply image enhancement" or "add human proofreading" – actions it cannot actually perform. Such suggestions are a form of capability hallucination, i.e. requests beyond the model’s abilities \cite{shi2024assessment}. Prior work shows that self-reflection generally improves model outputs \cite{shinn2023reflexion,yao2023react,madaan2023self}, but without constraints the model can propose invalid plans. We address this by filtering the chain-of-thought plan so that only model-executable steps remain. By restricting proposals to what the model can execute, capability reflection ensures each refinement step is realistic and grounded in the model’s actual capabilities.

Let the previous answer (or transcription) be $y_{\text{prev}}$ and let $\mathcal{P}=\{a_1,a_2,\ldots,a_n\}$ be the chain-of-thought plan proposed by the model in the reflection step. We define a feasibility indicator $\phi(a)$ that equals $1$ if action $a$ is within the model’s capability set and $0$ otherwise. Formally:

\[
\phi(a)=
\begin{cases}
1 & \text{if $a$ can be executed by the model} \\
  & \text{(e.g., text-based operations)} \\
0 & \text{if $a$ is infeasible (e.g., "image enhance")}
\end{cases}
\]

Using this indicator, we obtain the filtered plan
\[
\mathcal{P}_{\text{feas}} = \{\,a \in \mathcal{P} : \phi(a)=1\,\}.
\]

Finally, the model’s refinement mechanism $R$ (itself the LLM reasoning function) uses the feasible plan to produce an improved answer. If $x$ denotes the original input (e.g. the image and any context), the updated answer is
\[
y_{\text{new}} = R(x,\,y_{\text{prev}},\,\mathcal{P}_{\text{feas}}).
\]

Here $R(\cdot)$ denotes the model’s generation given the input, the previous answer, and only the filtered plan. In effect, we enforce
\[
\begin{aligned}
\mathcal{P}_{\text{feas}} &= \{a \in \mathcal{P} \mid a \text{ is within capability}\} \\
y_{\text{new}} &= R(x, y_{\text{prev}}, \mathcal{P}_{\text{feas}})
\end{aligned}
\]
ensuring the refinement only relies on feasible, model-plausible actions. This formalism cleanly captures the stepwise process of capability reflection.
\subsection{Memory Reflection}
We introduced memory reflection, designed to overcome the ``refinement stagnation'' and ``ineffective looping'' problems common in conventional iterative correction models. Unlike traditional models that may repeatedly attempt the same flawed strategies, our mechanism explicitly records and leverages a history of reflections. This ensures that each new refinement is informed by the entirety of past experiences, thus preventing redundant exploration of incorrect solution paths.

This process can be formalized as follows:

Let $I$ be the image input, $Q$ be the question, $A_i$ be the answer at iteration $i$, and $R_i$ be the reflection generated at iteration $i$. The initial answer, produced by zero-shot inference, is denoted as $A_0$.

We define the \textbf{Reflection Memory Store} $M_i$ as the set of all historical reflections prior to iteration $i$:
$$
M_i = \{R_1, R_2, \dots, R_{i-1}\}
$$

Within iteration $i$, the mechanism proceeds in the following steps:

\textbf{Reflection Generation:} Conditioned on the image $I$, question $Q$, the previous answer $A_{i-1}$, and the current memory store $M_i$, the model generates a new reflection $R_i$. This reflection aims to identify the inadequacies in $A_{i-1}$.
    $$
    R_i = \text{Reflect}(I, Q, A_{i-1}, M_i)
    $$

\textbf{Memory Update:} The newly generated reflection $R_i$ is incorporated into the memory store, creating an updated and more comprehensive historical record to guide the final refinement.
    $$
    M_{i+1} = M_i \cup \{R_i\}
    $$

 \textbf{Guided Refinement:} Instead of directly correcting $A_{i-1}$, the model generates a new, improved answer $A_i$ by conditioning on the original inputs and the \textbf{updated, complete memory store} $M_{i+1}$.
    $$
    A_i = \text{Refine}(I, Q, M_{i+1})
    $$

\begin{algorithm}[!htbp]
\caption{OCR-Agent with Capability and Memory Reflection}
\label{alg:refinement}
\begin{algorithmic}[1]
\REQUIRE Image $I$, Question $Q$, Initial answer $A_0$, Max iterations $T$
\ENSURE Final refined answer $A_T$

\STATE Initialize reflection memory: $M_1 \gets \emptyset$
\FOR{$i = 1$ \TO $T$}
    \STATE $R_i \gets \text{Reflect}(I, Q, A_{i-1}, M_i)$
    \STATE $\mathcal{P} \gets \text{ExtractPlan}(R_i)$  \COMMENT{Extract CoT plan from reflection}
    \STATE $\mathcal{P}_{\text{feas}} \gets \{ a \in \mathcal{P} \mid \phi(a) = 1 \}$  \COMMENT{$\phi(a)=1$ if model can execute $a$}
    \STATE $A_i \gets \text{Refine}(I, Q, A_{i-1}, \mathcal{P}_{\text{feas}}, M_i \cup \{R_i\})$
    \STATE $M_{i+1} \gets M_i \cup \{R_i\}$
\ENDFOR
\RETURN $A_T$
\end{algorithmic}
\end{algorithm}

\section{Experiments}
\subsection{Dataset and Metrics}
\paragraph{Datasets} To comprehensively and systematically evaluate the effectiveness of our proposed framework, we conducted a series of rigorous experiments on the comprehensive benchmark, OCRBench v2\cite{Fu2024OCRBenchVA}. This benchmark was chosen for its exceptional quality and comprehensive challenges: it contains over 10,000 manually verified question-answer pairs, covers both Chinese and English data, and includes a high proportion of difficult samples, enabling an effective assessment of the upper limits of a model's capabilities. Its task design extends beyond traditional text transcription to complex scenarios requiring deep visuotextual understanding, such as structured data extraction, visual question answering, and reasoning, which allows for a thorough examination of our framework's generalization ability across different languages and difficulty levels.

\paragraph{Metrics} For evaluation, we strictly adhere to its official standards, employing evaluation metrics tailored to six core task types. Specifically, for Parsing tasks, we use the Tree-Edit Distance Similarity (TEDS) to evaluate the quality of conversion to structured formats; for Localization tasks, the Intersection over Union (IoU) is used to measure the accuracy of predicted regions; for Extraction tasks, the F1 score is used to assess the performance of key information extraction and mapping; for Long Reading, we use a combination of BLEU, METEOR, F1 score, and Edit Distance for evaluation; for Counting tasks, we employ a normalized L1 distance to measure the accuracy of the counts; finally, for Basic VQA, we use Exact String Matching, a containment check, and the Average Normalized Levenshtein Similarity (ANLS) for evaluation, depending on the answer's length and type.

\begin{table*}[t]
\centering
\caption{Performance of VLMs on English subsets of OCRBench v2\cite{Fu2024OCRBenchVA}.}
\resizebox{\textwidth}{!}{
\begin{tabular}{lccccccccc}
\toprule
Method & Recognition & Referring & Spotting & Extraction & Parsing & Calculation & Understanding & Reasoning & Average \\
\midrule
\rowcolor{gray!20}
\multicolumn{10}{c}{\textbf{Open-source VLMs}} \\

LLaVA-Next-8B         & 41.3 & 18.8 & 0 & 49.5 & 21.2 & 17.3 & 55.2 & 48.9 & 31.5 \\
LLaVA-OV-7B           & 46.0 & 20.8 & 0.1 & 58.3 & 25.3 & 23.3 & 64.4 & 53.0 & 36.4 \\
Monkey-8B & 35.2 & 0 & 0 & 16.6 & 16.3 & 14.4 & 59.8 & 42.3 & 23.1 \\
TextMonkey-8B & 39.1 & 0.7 & 0 & 19.0 & 12.2 & 19.0 & 61.1 & 40.2 & 23.9 \\
Molmo-7B & 52.4 & 21.3 & 0.1 & 45.5 & 7.6 & 28.5 & 65.3 & 55.0 & 34.5 \\
Cambrian-1-8B & 45.3 & 21.5 & 0 & 53.6 & 19.2 & 19.5 & 63.5 & 55.5 & 34.7 \\
Pixtran-12B & 48.9 & 21.6 & 0 & 66.3 & 35.5 & 29.8 & 66.9 & 53.7 & 40.3 \\
InternVL3-8B  & 68.6 & 30.4 & 8.8 & \underline{85.3} & 34.0 & 27.1& \underline{77.5} & 60.3 & 49.0 \\
Deepseek-vl-2-Small-16B & 62.7 & 28.0 & 0.1 & 77.5 & 32.7 & 14.3 & 77.1 & 53.9 & 43.3 \\
MiniCPM-V-2.6-7B & 66.9 & 29.5 & 0.5 & 70.8 & 33.4 & 31.9 & 69.9 & 57.9 & 45.1 \\
GLM-4V-9B & 61.8 & 22.6 & 0 & 71.7 & 31.6 & 22.6 & 72.1 & 58.4 & 42.6 \\
Ovis2-8B & \underline{73.2} & 24.6 & 0.7 & 62.4 & \underline{44.8} & 40.6 & 72.7 & \underline{62.6} & 47.7 \\
RolmOCR-7B  &63.3 & 25.4 & 0.1 & 53.5 & 12.9 & 27.8 & 73.1 & 51.4 & 38.4 \\
\rowcolor{gray!20}
\multicolumn{10}{c}{\textbf{Closed-source VLMs}} \\
GPT-4o & 61.2 & 26.7 & 0 & 77.5 & 36.3 & 43.4 & 71.1 & 55.5 & 46.5 \\
GPT-4o-mini & 57.9 & 23.3 & 0.6 & 70.8 & 31.5 & 38.8 & 65.9 & 55.1 & 43.0 \\
Gemini-Pro & 61.2 & \underline{39.5} & \underline{13.5} & 79.3 & 39.2 & \underline{47.7} & 75.5 & 59.3 & \underline{51.9} \\
Claude3.5-sonnet & 62.2 & 28.4 & 1.3 & 56.6 & 37.8 & 40.8 & 73.5 & 60.9 & 45.2 \\
Step-1V & 67.8 & 31.3 & 7.2 & 73.6 & 37.2 & 27.8 & 69.8 & 58.6 & 46.7 \\
\midrule
\textbf{OCR-Agent-7B (Ours)}         & 71.8 & 27.7  & 0.7    & 80.8 & 39.4 & 41.3  & \textbf{79.9} & \textbf{66.5} & 51.0 \\
\bottomrule
\end{tabular}
}
\label{tab:results eng}
\end{table*}

\begin{figure*}[t]
    \centering
    \begin{subfigure}
        \centering
        \includegraphics[width=0.24\textwidth]{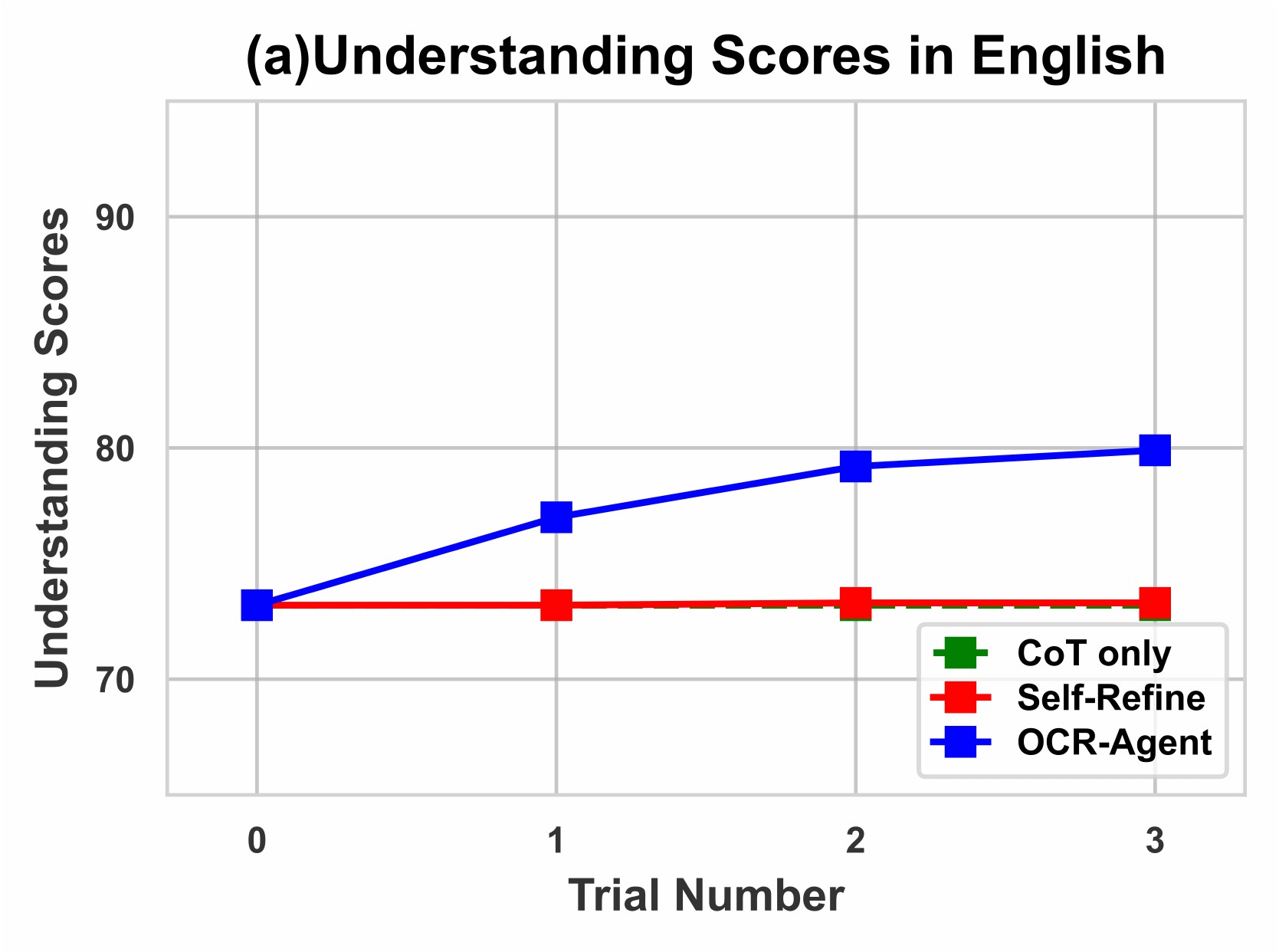}
    \end{subfigure}
    \begin{subfigure}
        \centering
        \includegraphics[width=0.24\textwidth]{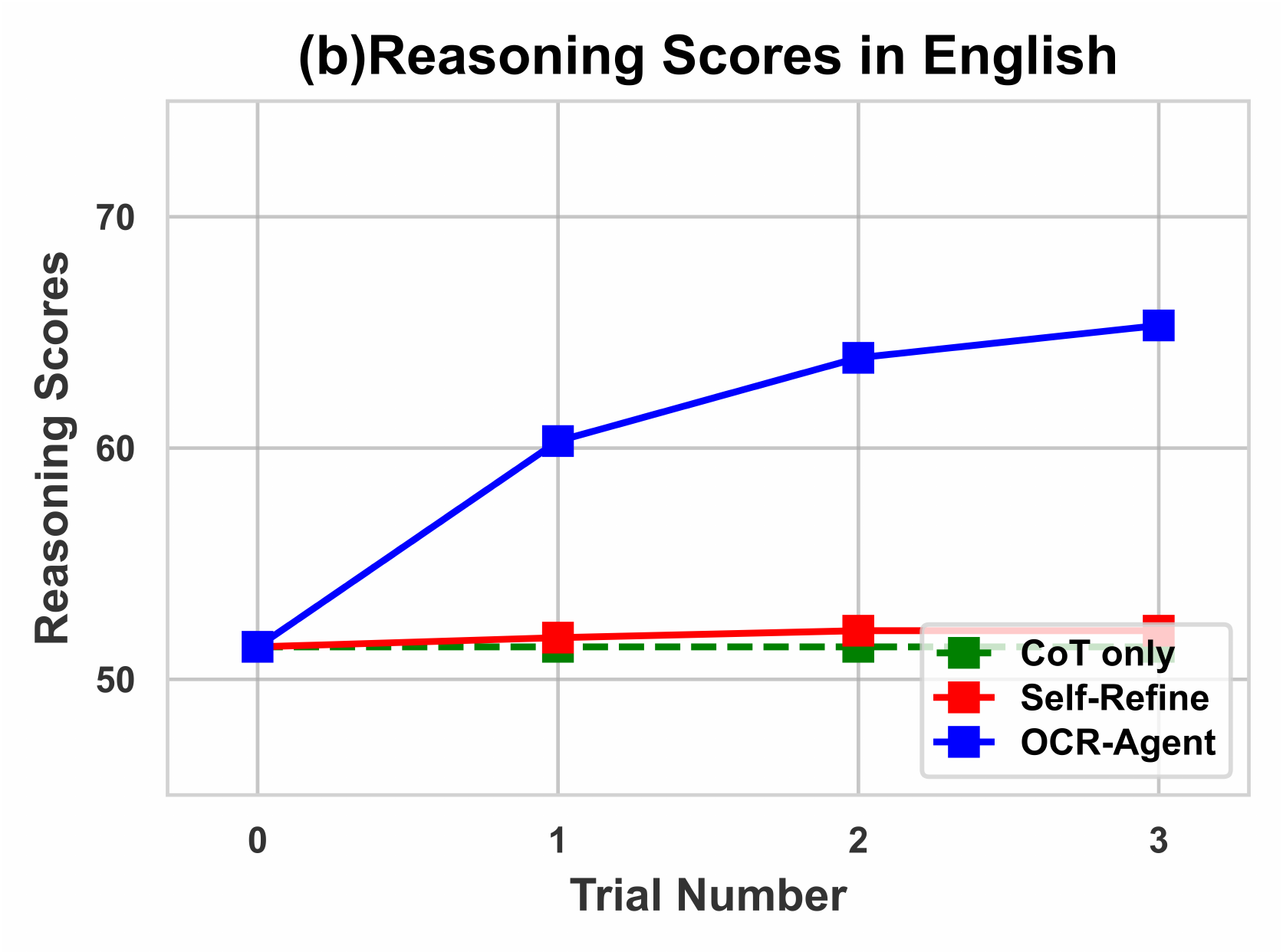}
        \label{fig:reasoning_en_curve}
    \end{subfigure}
    \begin{subfigure}
        \centering
        \includegraphics[width=0.24\textwidth]{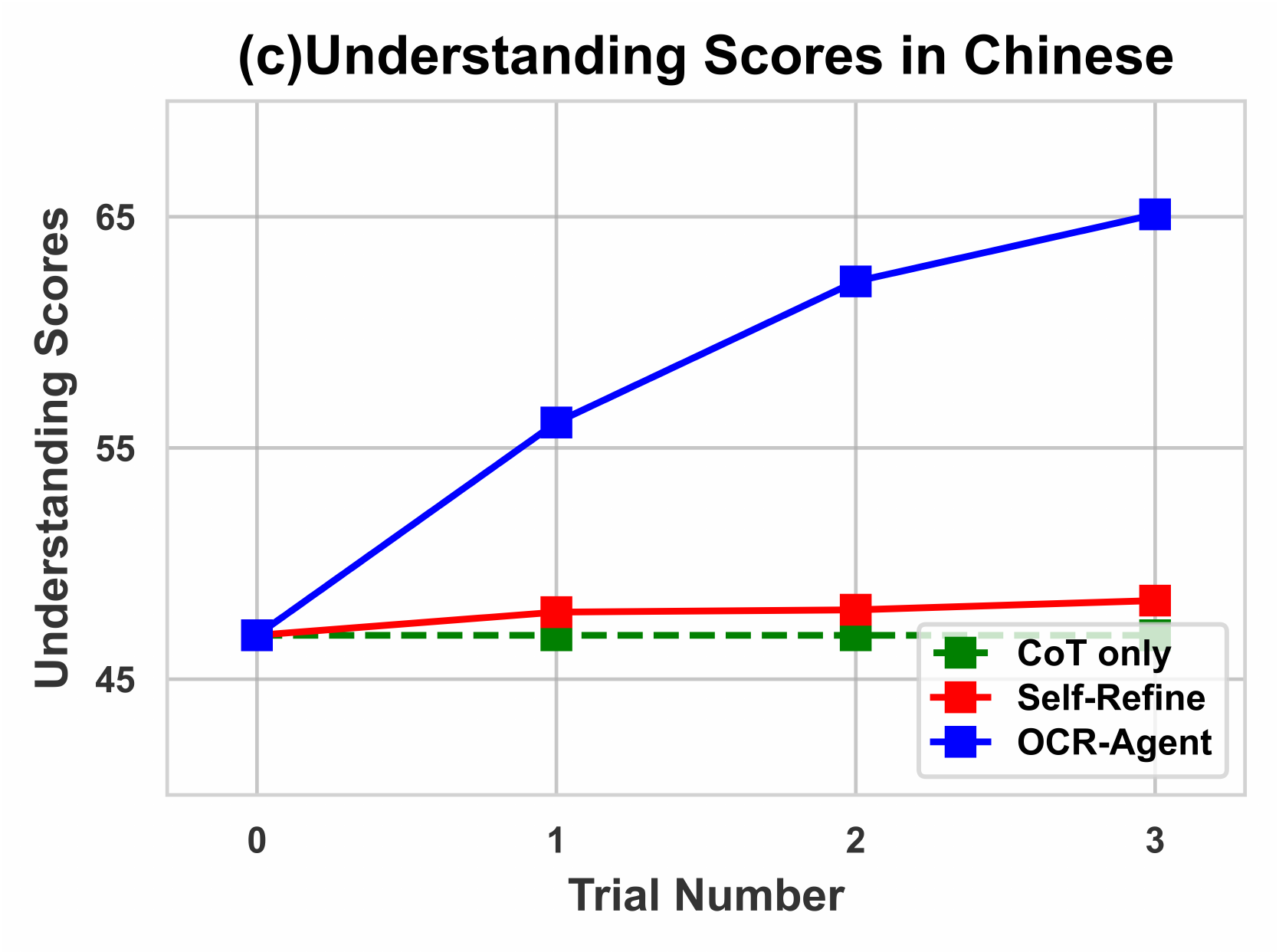}
        \label{fig:understanding_cn_curve}
    \end{subfigure}
    \begin{subfigure}
        \centering
        \includegraphics[width=0.24\textwidth]{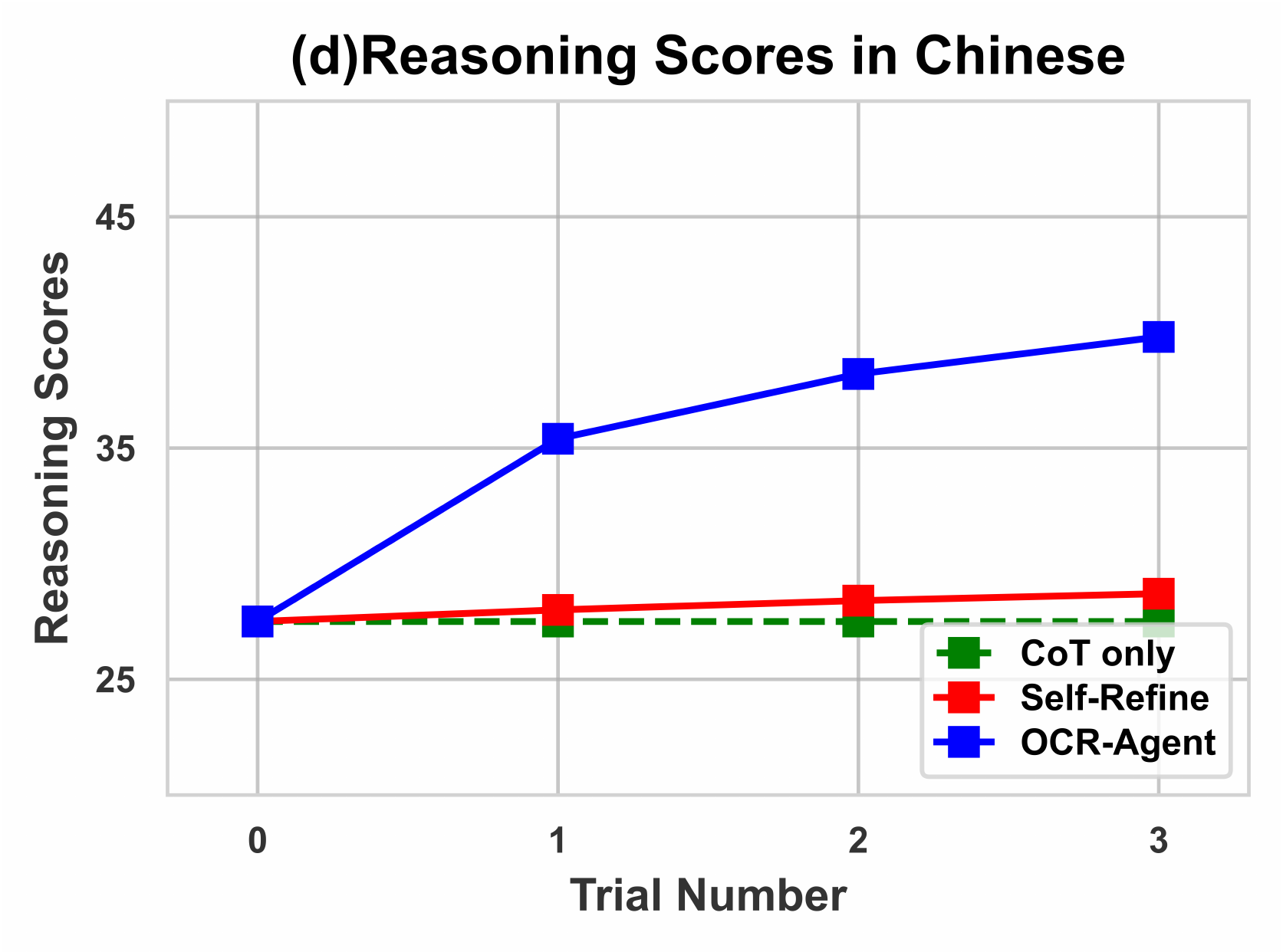}
    \end{subfigure}
    \caption{ The performance improvement of OCR-Agent in understanding and reasoning as the number of trials increases. \textbf{(a)\&(c)} Understanding Scores in English and Chinese: It depicts how the understanding capabilities of CoT only, Self-Refine, and OCR-Agent evolve with increasing trial numbers. \textbf{(b)\&(d)} Reasoning Scores in English and Chinese: It shows the reasoning ability changes of these methods as trial numbers proceed.}
     \label{fig:capability curve with trial number}
\end{figure*}

\subsection{Implementation Details}
For each sample in the OCRBench v2 dataset, our method first performs a standard, simple inference to generate an initial baseline answer. Subsequently, the system enters an iterative refinement process, which is fixed for three rounds (N=3). In each iteration i, this process involves two core stages: 1. Reflection Generation: We utilize a "reflection" prompt template, which integrates the original question, the answer from the previous iteration (i-1), and the accumulated history of reflections, to have the model generate new reflective text. 2. Guided Refinement: We then feed the original question and the updated, complete history of reflections into a "refinement" prompt template to guide the model in generating the revised answer for iteration i. This answer serves as the input for the next iteration, thereby enabling progressive refinement. The entire inference process was executed on four NVIDIA 3090 GPUs.
\subsection{Main Results}
We conduct a comprehensive evaluation of the proposed OCR-Agent framework on the highly challenging OCRBench v2 benchmark, covering both English and Chinese scenarios across eight core tasks: Recognition, Referring, Spotting, Extraction, Parsing, Calculation, Understanding, and Reasoning. As shown in Tables~\ref{tab:results eng} and~\ref{tab:results china}, our method significantly outperforms current mainstream open-source and closed-source Vision-Language Models (VLMs) — without requiring any additional training.

On the English subset, OCR-Agent achieves an average score of 51.01, surpassing all open-source models and closely approaching the strongest closed-source model, Gemini-Pro (51.9). Notably, it achieves the highest scores on the two most challenging tasks — Visual Understanding (79.9) and Visual Reasoning (66.5) — demonstrating the powerful advantage of our dual-reflection mechanism in handling complex multimodal reasoning.

On the Chinese subset, OCR-Agent also delivers outstanding performance, achieving an average score of 54.72, ranking second only to the current top-performing open-source model, Qwen2.5-VL-7B (55.6). It sets new open-source records in Text Recognition (77.0), Information Extraction (68.8), and Visual Understanding (65.1). Importantly, while the base RolmOCR-7B model scores only 38.6 on Chinese tasks, OCR-Agent boosts its performance by nearly 16 points, highlighting the framework’s strong generalization and enhancement capability across diverse base models.

Overall, OCR-Agent despite its relatively lightweight size (7B parameters) — outperforms larger models such as Pixtral-12B and Deepseek-VL2-16B on multiple key tasks. This convincingly validates the effectiveness of our Capability Reflection and Memory Reflection mechanisms in guiding efficient, grounded, and iterative self-correction.

\begin{table*}[t]
\centering
\caption{Performance of VLMs on Chinese subsets of OCRBench v2\cite{Fu2024OCRBenchVA}.}
\resizebox{0.75\textwidth}{!}{
\begin{tabular}{lcccccc}
\toprule
Method & Recognition   & Extraction & Parsing  & Understanding & Reasoning & Average \\
\midrule
\rowcolor{gray!20}

\multicolumn{7}{c}{\textbf{Open-source VLMs}} \\

LLaVA-Next-8B  & 5.7 & 2.9 & 12.2 & 7.5 & 17.2 & 9.1 \\
LLaVA-OV-7B  & 14.8 & 15.7 & 13.7 & 16.0 & 28.7 & 17.8 \\
Monkey-8B &4.6  & 11.2 & 8.4 & 21.5 & 20.0 & 13.1 \\
TextMonkey-8B  & 23.5 & 14.8 & 8.4 & 19.9 & 12.2 & 15.8 \\
Molmo-7B  & 7.1 & 15.0 & 9.2 & 9.0 & 23.7 & 12.8 \\
Cambrian-1-8B  & 5.3 & 14.9 & 12.6 & 8.5 & 8.1 & 9.9 \\
Pixtral-12B  & 13.4 & 10.9 & 21.0 & 7.0 & 20.7 & 14.6 \\
InternVL3-8B  & 68.9 & 62.0 & 31.6 & \underline{57.9}  & \underline{47.3} & \underline{53.5}\\
Deepseek-VL2-Small-16B  & 60.9 & 50.6 & 28.3 & 53.0 & 20.5 & 42.7 \\
MiniCPM-o-2.6-7B  & 53.0 & 49.4 & 27.1 & 43.5 & 32.7 & 41.1 \\
GLM-4V-9B  & 24.4 & 60.6 & 20.4 & 52.8 & 25.2 & 36.6 \\
Ovis2-8B  & \underline{72.2} & 50.8 & 37.7 & 47.9 & 37.4 & 49.2 \\
RolmOCR-7B  & 36.5 & \underline{64.9} & 15.3 & 45.4 & 26.7 & 37.7\\
\rowcolor{gray!20}
\multicolumn{7}{c}{\textbf{Closed-source VLMs}} \\
GPT-4o  & 21.6 & 53.0 & 29.8 & 38.5 & 18.2 & 32.2 \\
GPT-4o-mini  & 13.1 & 38.9 & 27.2 & 28.8 & 16.9 & 25.0 \\
Gemini-Pro  & 52.5 & 47.3 & 30.9 & 51.5 & 33.4 & 43.1 \\
Claude3.5-sonnet  & 21.0 & 56.2 & 35.2 & 55.0 & 30.5 & 39.6 \\
Step-1V  & 56.7 & 41.1 & \underline{37.6} & 38.3 & 39.2 & 42.6 \\
\midrule
\textbf{OCR-Agent-7B (Ours)}  & \textbf{77.0}  & \textbf{69.1} & 22.9  & \textbf{65.1}  & 39.8 & \textbf{54.7} \\
\bottomrule
\end{tabular}
}
\label{tab:results china}
\end{table*}

\begin{table*}[t]
\centering
\caption{Comparison of different methods (including naive RolmOCR and self-reflection variants) on English subsets of OCRBench v2.}
\resizebox{\textwidth}{!}{
\begin{tabular}{lccccccccc}
\toprule
Method & Recognition & Referring & Spotting & Extraction & Parsing & Calculation & Understanding & Reasoning & Average \\
\midrule
Naive & 63.3 & 25.4 & 0.1 & 53.5 & 12.9 & 27.8 & 73.1 & 51.4 & 38.4 \\
CoT & 64.5 & 25.1 & 0.1 & 76.5 & 17.8 & 31.9 & 73.2 & 51.9 & 42.0 \\
Self-Refine & 65.4 & 25.3 & 0.1 & 78.3 & 19.4 & 33.3 & 73.3 & 52.1 & 43.4 \\
Capability Reflection & 67.5 & 26.1 & 0.3 & 79.1 & 26.1 & 36.0 & 75.5 & 56.9 & 45.9 \\
Memory Reflection & 69.6 & 26.9 & 0.5 & 79.9 & 32.7 & 38.6 & 77.7 & 61.7 & 48.4 \\
Capability \& Memory & \underline{71.8} & \underline{27.7}  & \underline{0.7}   & \underline{80.8} & \underline{39.4} & \underline{41.3}  & \underline{79.9} & \underline{66.5} & \underline{51.0} \\
\bottomrule
\end{tabular}
}
\label{tab:results_english}
\end{table*}

\subsection{Ablation Study}
To systematically evaluate the contribution of each component, we conduct a detailed ablation study of OCR-Agent on both the English and Chinese subsets of OCRBench v2. All variants are compared against the standard Self-Refine baseline. As summarized in Tables~\ref{tab:results_english} and~\ref{tab:results_chinese}, OCR-Agent achieves consistent and stable performance gains across both Chinese and English tasks, with particularly pronounced improvement on Understanding and Reasoning tasks as the number of iteration rounds increases.

As visualized in Fig~\ref{fig:capability curve with trial number}, the performance curves clearly demonstrate that while baseline methods (CoT, Self-Refine) plateau or fluctuate after the first or second iteration, OCR-Agent continues to improve steadily across all three rounds — especially in high-complexity tasks such as English Reasoning (Fig~\ref{fig:reasoning_en_curve}) and Chinese Understanding (Fig~\ref{fig:understanding_cn_curve}). This sustained progression validates the effectiveness of our dual-reflection design in avoiding stagnation and enabling deep iterative refinement.

When Capability Reflection and Memory Reflection are combined, the full OCR-Agent achieves peak performance, with average scores of 51.0 (English) and 54.7 (Chinese), demonstrating a clear complementary effect. Notably, on the Chinese Recognition task, performance surges from 37.7 to 77.0, highlighting the framework’s strong task adaptability.

In summary, Capability Reflection ensures the feasibility of each refinement step, while Memory Reflection enables progressive exploration across iterations. Together, they form an iterative optimization framework that significantly outperforms traditional self-refinement methods.

\begin{table*}[t]
\centering
\caption{Comparison of different methods (including naive RolmOCR and self-reflection variants) on Chinese subsets of OCRBench v2.}
\resizebox{0.75\textwidth}{!}{
\begin{tabular}{lcccccc}
\toprule
Method & Recognition   & Extraction & Parsing  & Understanding & Reasoning & Average \\
\midrule
Naive  & 36.5 & 64.9 & 15.3 & 45.4 & 26.7 & 37.7 \\
CoT & 37.1 & 66.5 & 14.1 & 46.9 & 27.5 & 38.4 \\
Self-Refine  & 37.7 & 68.8 & 13.7 & 48.4 & 28.7 & 39.4 \\
Capability Reflection & 50.8 & 68.9 & 16.8 & 54.0 & 32.4 & 44.6 \\
Memory Reflection & 63.9 & 69.0 & 19.8 & 59.5 & 36.1 & 49.7 \\
Capability \& Memory & \underline{77.0}  & \underline{69.1} & \underline{22.9}  & \underline{65.1}  & \underline{39.8} & \underline{54.7} \\
\bottomrule
\end{tabular}
}
\label{tab:results_chinese}
\end{table*}

\section{Limitation and Future Work}

While OCR-Agent demonstrates significant performance improvements through its novel self-correction framework, several limitations remain, pointing towards promising directions for future research.

\textbf{Computational Overhead}: The iterative reflection and refinement process inherently requires multiple calls to the large VLM for a single input, increasing inference time and computational cost compared to single-pass models. This could hinder deployment in real-time applications. Our three-round iteration scheme, while effective, may also be inefficient for simpler problems that require fewer steps and redundant for extremely complex ones that need more.

\textbf{Dependence on Base Model Capabilities}: The framework's effectiveness is ultimately bounded by the inherent capabilities of the base VLM (e.g., RolmOCR). If the base model fundamentally misperceives a critical visual element or lacks specific knowledge, the reflection process may be unable to recover from this initial error, leading to refinement within a wrong context.

Future directions include optimizing the framework’s efficiency through dynamic iteration control and model distillation, integrating external tools (e.g., image super-resolution APIs) to overcome inherent model limitations, and extending its application to broader vision-language tasks such as chart understanding and multimodal reasoning. Further enhancements to the memory mechanism, such as structured knowledge storage, and exploring human-in-the-loop refinement for high-stakes scenarios, are also promising avenues.

\section{Conclusion}

In this paper, we identified that unconstrained self-reflection often leads to unstable reasoning and ineffective corrections, ultimately impairing model performance.

To overcome these limitations, we proposed \textbf{OCR-Agent}, a novel reflection-based framework that incorporates two core mechanisms: \textbf{Capability Reflection}, and \textbf{Memory Reflection}. Together, these components enable more structured and sustainable self-correction, leading to consistent performance improvements without additional fine-tuning.

Extensive experiments on the OCRBench v2 benchmark demonstrate that our approach significantly outperforms both direct answering, standard CoT prompting, and simple self-refine strategies, confirming its effectiveness in enhancing VLM robustness in text-rich visual understanding tasks.

Our findings affirm that carefully constrained self-reflection can unlock more robust and sustainable reasoning in VLMs, paving the way for more reliable and interpretable multimodal systems. Future work will focus on optimizing the computational efficiency of the reflection process and extending the framework to a broader range of vision-language tasks.

\clearpage
\clearpage
\bibliographystyle{IEEEbib}
\bibliography{bare_conf_compsoc}

\end{document}